\documentclass[letterpaper]{article} 
\usepackage{aaai24}  
\usepackage{times}  
\usepackage{helvet}  
\usepackage{courier}  
\usepackage[hyphens]{url}  
\usepackage{graphicx} 
\usepackage{natbib}  
\usepackage{caption} 
\usepackage{amsmath}
\usepackage{booktabs}       
\usepackage{amsfonts}       
\usepackage{nicefrac}       
\usepackage{microtype}      
\usepackage{xcolor}         
\usepackage{lipsum}

\usepackage{graphicx}
\usepackage{multirow}
\usepackage[capitalize,noabbrev]{cleveref}
\usepackage{algorithm}
\usepackage{algpseudocode}

\usepackage{newfloat}
\usepackage{listings}

\DeclareCaptionStyle{ruled}{labelfont=normalfont,labelsep=colon,strut=off} 
\floatstyle{ruled}
\newfloat{listing}{tb}{lst}{}
\floatname{listing}{Listing}
\title{Out of Thin Air: Exploring Data-Free Adversarial Robustness Distillation}
\author{
Yuzheng Wang\textsuperscript{\rm 1}\thanks{Equal contributions} $\quad$
Zhaoyu Chen\textsuperscript{\rm 1}$^{*}$ $\quad$
Dingkang Yang\textsuperscript{\rm 1} $\quad$
Pinxue Guo\textsuperscript{\rm 1} \\
Kaixun Jiang\textsuperscript{\rm 1} $\quad$
Wenqiang Zhang\textsuperscript{\rm 1,2}  $\quad$
Lizhe Qi\textsuperscript{\rm 1}\thanks{Corresponding author}\\
}

\affiliations{
    \textsuperscript{\rm 1}Shanghai Engineering Research Center of AI \& Robotics, Academy for Engineering \& Technology, Fudan University \\
    \textsuperscript{\rm 2}Engineering Research Center of AI \& Robotics, Ministry of Education, Academy for Engineering \& Technology, Fudan University \\
    \tt\small{\{yzwang20, zhaoyuchen20\}@fudan.edu.cn}

}

\usepackage{bibentry}

\begin{document}

\maketitle

\begin{abstract}

Adversarial Robustness Distillation (ARD) is a promising task to solve the issue of limited adversarial robustness of small capacity models while optimizing the expensive computational costs of Adversarial Training (AT).
Despite the good robust performance, the existing ARD methods are still impractical to deploy in natural high-security scenes due to these methods rely entirely on original or publicly available data with a similar distribution.
In fact, these data are almost always private, specific, and distinctive for scenes that require high robustness.
To tackle these issues, we propose a challenging but significant task called Data-Free Adversarial Robustness Distillation (DFARD), which aims to train small, easily deployable, robust models without relying on data.
We demonstrate that the challenge lies in the lower upper bound of knowledge transfer information, making it crucial to mining and transferring knowledge more efficiently.
Inspired by human education, we design a plug-and-play Interactive Temperature Adjustment (ITA) strategy to improve the efficiency of knowledge transfer and propose an Adaptive Generator Balance (AGB) module to retain more data information.
Our method uses adaptive hyperparameters to avoid a large number of parameter tuning, which significantly outperforms the combination of existing techniques.
Meanwhile, our method achieves stable and reliable performance on multiple benchmarks.

\end{abstract}

\section{Introduction}

Deep learning has achieved great success in many fields \cite{devlin2018bert,dosovitskiy2020image,yang2023target,yang2023context,yang2023aide,yang2023how2comm,liu2023amp,liu2023learning,liu2023improving,wang2023explicit,wang2023sampling}.
Along with this process, deep learning models are increasingly expected to be deployed in established and emerging artificial intelligence fields.
However, high-performance models' large scale and high computational costs \cite{ramesh2022hierarchical} prevent this technology from being applied to mobile devices, driverless cars, and tiny robots.
More importantly, many studies have shown that well-trained deep learning models are vulnerable to adversarial examples containing only minor changes \cite{goodfellow2014explaining,chen2022shape,chen2023content}.
Therefore, training robust small-capacity models has become the key to breaking the bottleneck.

Various defensive strategies have been proposed \cite{madry2017towards,jia2019comdefend,chen2022towards,wang2023adversarial} for adversarial robustness.
Among them, Adversarial Training (AT) has been considered the most effective approach \cite{athalye2018obfuscated,croce2020reliable}.
By generating adversarial examples, the models can learn robustness knowledge to deal with various adversarial attacks.
Therefore, it can significantly improve the robustness of large-capacity models.
However, the robustness performance is struggling for small models only relying on AT due to the limited model capacity.
Based on this, guided by the pre-trained robust teacher with insights, the robustness of small models is improved.
This process is called Adversarial Robustness Distillation (ARD) \cite{goldblum2020adversarially}.

\begin{figure*}[t]
	\centering
	\includegraphics[scale=0.29]{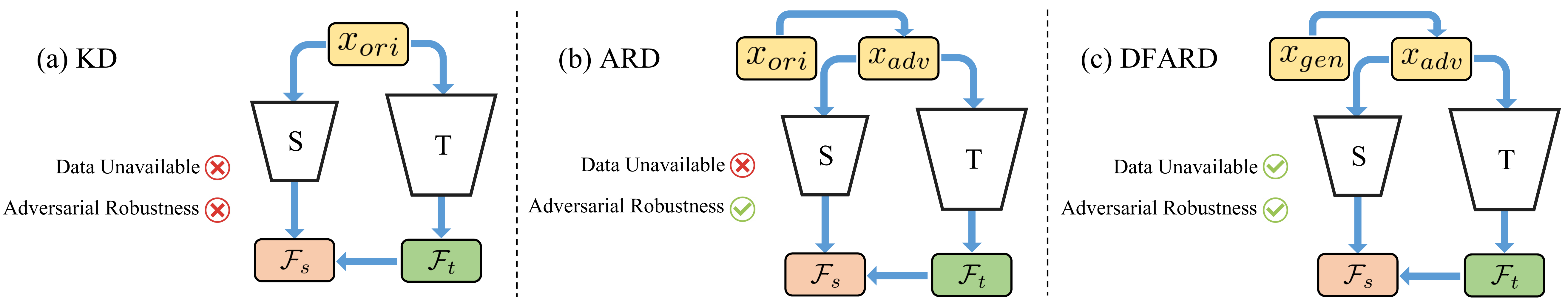}
    \vspace{-0.1cm}
	\caption{Diagrams of (a) Knowledge Distillation (KD), (b) Adversarial Robustness Distillation (ARD), and (c) Data-Free Adversarial Robustness Distillation (DFARD). \textbf{S} and \textbf{T} represent the student and the teacher network respectively. $\mathcal{F}_{s}$ and $\mathcal{F}_{t}$ represent the search spaces. $x_{ori}$ and $x_{gen}$ is the original and generated data. $x_{adv}$ is the adversarial examples.}
    \vspace{-0.1cm}
	\label{fig1}
\end{figure*}

Despite improving the robustness of small models, existing ARD methods are still hard to apply in real-world scenes due to impractical settings.
The original training data is of primary concern.
Firstly, all existing ARD methods assume that the original training data is available throughout the distillation process \cite{goldblum2020adversarially,zhao2022enhanced}.
In practical application, for scenes with high robustness requirements, the original data is usually private and unavailable (\textit{e.g.}, face data for face recognition system, disease data for medical diagnosis, and financial data for quantitative investment).
Secondly, some technologies avoid relying on original private data by using open-world or out-of-domain (OOD) unlabeled data \cite{fang2021mosaicking}.
However, these methods rely on a necessary assumption that private data can always be obtained simply from open datasets.
Although some methods claim to use OOD data, the performance of these methods degrades drastically when the discrepancy between the unavailable original data and unlabeled data increases.
Therefore, good performance extremely depends on broadly similar image patterns between the data domains instead of OOD \cite{yang2021generalized}.
Based on these, existing technologies are still challenging to deploy in high-security robustness scenes.
\textbf{One question is whether we can efficiently train small, easily deployable, robust models to improve robustness without original private data and specific data with similar patterns}.
To explore this question, we propose a novel task called Data-Free Adversarial Robustness Distillation (DFARD).
The diagrams are shown in Figure \ref{fig1}.
Compared with the existing KD (a) and ARD (b) tasks, our DFARD only uses generated data, which is more general and practical.

Considering the knowledge transfer process between the teacher and student networks, we demonstrate that the information upper bound is lower in the DFARD than in existing tasks.
While removing the ARD task's dependence on private data, the challenges lie in less effective knowledge transfer and less data knowledge in the generated data.

To tackle the issues, we select the commonly used generator training objectives as a DFARD baseline and optimize it from the following aspects:
\textbf{1)} To improve the effectiveness of knowledge transfer, we first propose an Interactive Temperature Adjustment (ITA) strategy to help students find more suitable training objectives for each training epoch.
\textbf{2)} To retain more data information, we then propose an Adaptive Generator Balance (AGB) module to better balance the similarity of the data domains and the information content.
In addition, our method uses adaptive hyperparameters to avoid a large number of parameter tuning.
Specifically, the primary contributions and experiments are summarized below:
\begin{itemize}
    \item To our best knowledge, we are the first to propose a novel task named DFARD to apply higher security level application scenes. Further, we theoretically demonstrate the challenges of this new task via the information bound.
    \item We optimize DFARD to improve the effectiveness of knowledge transfer and retain more data information.
    A plug-and-play ITA strategy and an AGB module are proposed to gain the simplest combination of generator losses, avoiding complex loss designs and weight balance, significantly reducing parameter tuning costs.
    \item Experiments show that our DFARD method achieves stable and reliable performance on multiple benchmarks comparing combinations of existing technologies.
\end{itemize}

\section{Related Work}

\subsection{Data-Free Generation}

Data-free generation technology is proposed to generate substitute data with Generative Adversarial Networks (GANs) or other generation modules.
During this process, researchers do not need to access any data, thus being able to deal with data privacy and other data unavailable issues.
Chen \textit{et al.} \cite{chen2019data} first introduce the generator into a data-free generation process to get more vital generation capabilities.
To obtain the generated data that the student does not learn well, Micaelli \textit{et al.} \cite{micaelli2019zero} introduce the method of adversarial generation.
They prompt the generator to generate data with more significant differences between the student's and teacher's predictions so that the shortcomings are made up in the learning process.
Choi \textit{et al.} \cite{choi2020data} add batch categorical entropy into the data-free generation process to promote class balance.
To further improve the generation speed, Fang \textit{et al.} \cite{fang2022up} propose feature sharing to simplify the generation process of each step.
To improve generation quality, Bhardwaj \textit{et al.} \cite{bhardwaj2019dream} introduce model inversion and use the intermediate layer statistics of the teacher model to restore the original data.
Based on this, Yin \textit{et al.} \cite{yin2020dreaming} introduce adversarial inversion, and Fang \textit{et al.} \cite{fang2021contrastive} introduce contrastive learning to enhance the generation quality further.

\begin{figure*}[t]
	\centering
	\includegraphics[scale=0.37]{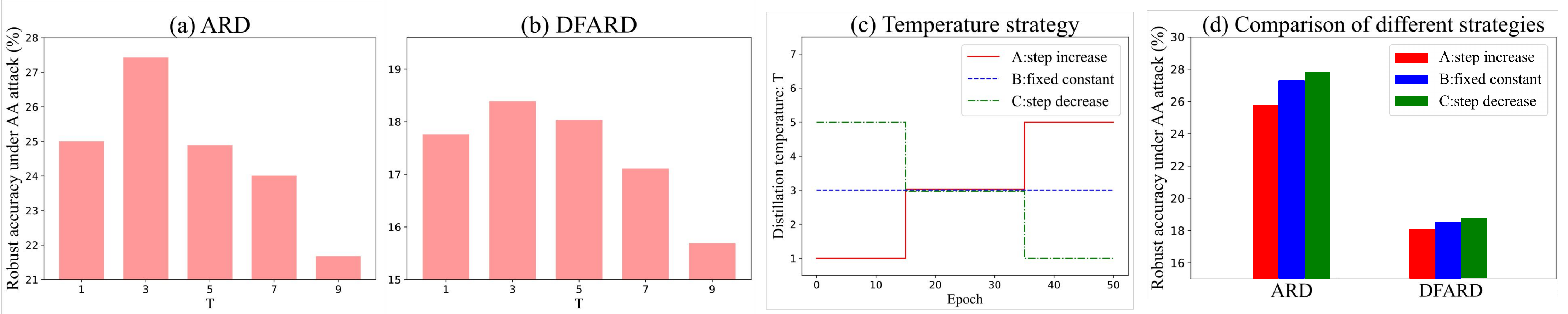}
    \vspace{-0.2cm}
	\caption{A toy experiment about the effect of different temperatures and simple verification of the easy-to-hard process. (a) and (b) show the impact of using different fixed temperatures for student performance on ARD and DFARD, respectively. (c) shows simple step temperature strategies, which means the trend of varying difficulty changes in the learning objective. (d) shows the performance comparison of students under these strategies.}
	\label{fig3}
    \vspace{-0.1cm}
\end{figure*}

\subsection{Adversarial Robustness Distillation}

Early adversarial training methods focus on learning directly from adversarial examples to improve model adversarial robustness \cite{madry2017towards,zhang2019theoretically}.
However, expanding the adversarial training set leads to increased training costs.
More importantly, the robustness improvement of small models is not evident due to the limitation of model capacity.
Adversarial robustness distillation is proposed to address these issues.
The setting is that both pre-trained robust teacher models and original training data are available.
Goldblum \textit{et al.} \cite{goldblum2020adversarially} first propose the concept of adversarial robustness distillation.
They show that improving the robustness of small models is feasible without additional training costs.
Zi \textit{et al.} \cite{zi2021revisiting} find that the soft labels given by the teacher are very effective and can significantly improve the robustness performance of the student.
Zhu \textit{et al.} \cite{zhu2022reliable} find that the teacher's confidence in the student's adversarial examples continues to decline, which may not be able to give the correct guidance.
They propose a multi-stage strategy to allow the student to learn independently in later training.
Zhao \textit{et al.} \cite{zhao2022enhanced} utilize multiple teachers to learn from nature and robust scenes separately.
Based on this, they try to focus on clean accuracy while improving adversarial robustness.

\section{The Challenges of DFARD}

To explore the impact of missing original training data on existing ARD tasks, we start with the effectiveness of knowledge transfer in the distillation process.
By analyzing the lipschitzness of the robust model and the properties of the generated data, we theoretically demonstrate why DFARD is more challenging than KD and ARD tasks.
DFARD has a lower information upper limit than KD and ARD in the knowledge transfer process.
This conclusion implies that for the DFARD task, more knowledge is needed.
Based on this, we try to improve the efficiency of knowledge transfer and ensure higher data information to meet this challenge.
Inspired by Human Education and Curriculum Learning \cite{bengio2009curriculum,pentina2015curriculum}, we try to look at the knowledge transfer process from the perspective of \textbf{1)} the knowledge from the teacher and \textbf{2)} the knowledge from the data.
Detailed discussions are as follows:

\begin{figure}[h]
	\centering
	\includegraphics[scale=0.4]{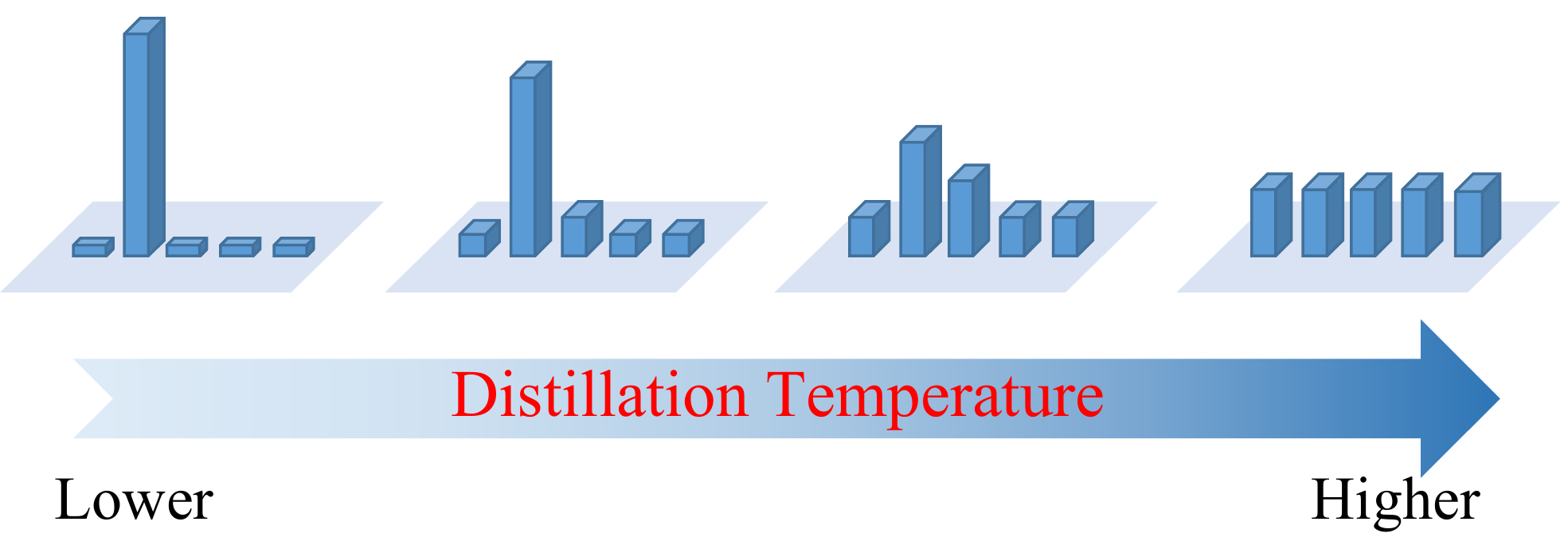}
    \vspace{-0.2cm}
	\caption{Diagrams of teacher's soft labels. As the temperature increases, the distance between the teacher's and naive student's prediction distributions decreases, and the difficulty of the learning objectives decreases.}
	\label{fig2}
    \vspace{-0.1cm}
\end{figure}

\noindent \textbf{The Knowledge from the Teacher.}
In the process of human education, teachers always teach students simple knowledge in the beginning.
With improving students' abilities, more difficult knowledge is gradually covered.
This easy-to-hard training process improves the efficiency of knowledge transfer.
Inspired by this, we analogize the DFARD task to the complex challenge of the human learning process.
The key lies in how to build an easy-to-hard process.
In fact, distillation temperature enables the teacher network to provide suitable soft labels to transfer knowledge from the cumbersome model to a small model \cite{hinton2015distilling, romero2014fitnets}.
The temperature controls the discrepancy between two distributions and represents learning objectives of varying degrees of difficulty \cite{muller2019does, liasymmetric, zi2021revisiting,li2022curriculum} as shown in Figure \ref{fig2}.
Most existing methods ignore the usefulness of the distillation temperature itself, regard it as a fixed hyperparameter, and inefficiently search for optimum.
On this basis, they spend several times on computational costs.

To verify that the easy-to-hard process can or cannot improve knowledge transfer efficiency and better deal with DFARD tasks, we conduct a toy experiment as shown in Figure \ref{fig3}.
We first verify the effect of different fixed distillation temperatures on two tasks.
We train all student models for 50 epochs with or without original data and report the best robustness accuracy under AutoAttack (AA) attack \cite{croce2020reliable}.
From Figure \ref{fig3}(a) and (b), a general conclusion is that different temperatures have effects on the two tasks.
Further, we respectively construct three temperature strategies of step increase, fixed constant, and step decrease to build learning objectives with different difficulties for each epoch (Figure \ref{fig3}(c)).
The inflection points of temperature change are at the 15th and 35th epochs.
Based on these strategies, we test the robustness performance as shown in Figure~\ref{fig3}(d).
We find that the strategy of step decrease achieves the best results.
As shown in Figure~\ref{fig2}, the decrease in temperature means the learning difficulty increases.
That is, the easy-to-hard knowledge promotes the student's progress.

\begin{figure*}[t]
	\centering
	\includegraphics[scale=0.39]{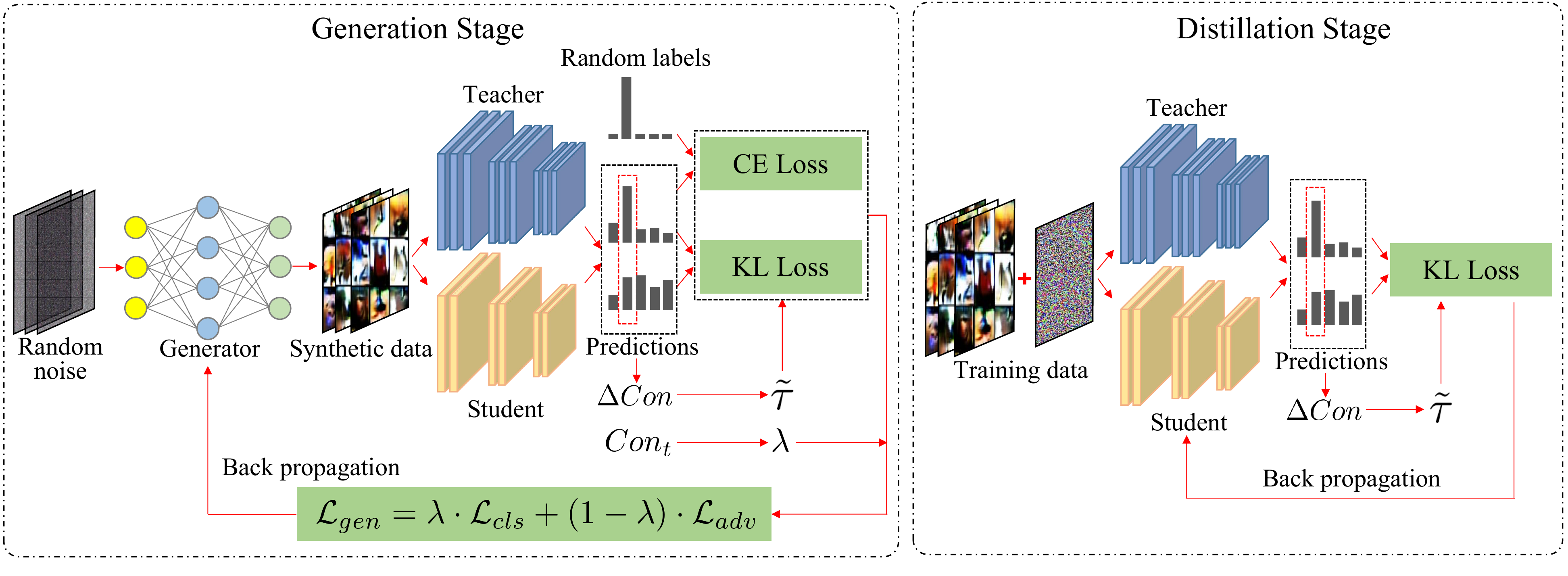}
    \vspace{-0.2cm}
	\caption{The pipeline of our optimized DFARD method from the most commonly used training baseline. Our method consists of two stages: (1) In the generation stage, we design an Interactive Temperature Adjustment strategy to adjust the temperature $\tilde{\tau}$ according to the student's learning. Simultaneously, we propose an Adaptive Generator Balance module to balance the similarity between data domains and the information content of data. (2) In the distillation stage, we keep the interactive temperature to help the student learn better. Training data represents the generated adversarial examples.}
	\label{fig4}
    \vspace{-0.1cm}
\end{figure*}

\noindent \textbf{The Knowledge from the Data.}
For human education, apart from teachers, good tutorials are equally important.
Generally speaking, tutorials that contain more knowledge can give students more help in the process of learning.
More knowledge with more information is crucial.
Inspired by this, we analogize the generated data as a medium of knowledge transfer to the tutorials.
The generated data with different information content may also help students differently.
Existing all data-free generation methods set a fixed generation loss weight to train a generator and constrain the teacher's confidence in the generated data \cite{chen2019data,yin2020dreaming,choi2020data,fang2021contrastive,fang2022up}.
To obtain generated data that is closer to the original distribution, the teacher model's predictions $f_{t}(\hat{\textbf{\textrm{x}}})$ should be close to the one-hot labels $\textbf{\textrm{y}}$, \textit{e.g.}, minimize them with the following cross-entropy loss:
\begin{equation}\label{eq1}
\mathcal{L}_{cls}=CE(f_{t}(\hat{\textbf{\textrm{x}}}), \textbf{\textrm{y}}),
\end{equation}
where $\textbf{\textrm{y}}$ can be randomly generated labels or pseudo-labels of the teacher.
$\hat{\textbf{\textrm{x}}}$ is synthesized by the generator $g$ through random noise $\textbf{\textrm{z}}$ and the label $\textbf{\textrm{y}}$: $\hat{\textbf{\textrm{x}}}=g(\textbf{\textrm{z}},\textbf{\textrm{y}})$.

In the above process, the teacher provides a more prominent target logit or less varied wrong logits.
We argue that the above process gradually decreases the information content of the teacher's soft labels.
The information content is a basic quantity derived from the probability prediction for the generated data of the teacher model.
In this paper, we measure the information content by Information Entropy.
Some studies have shown that such soft labels reduce information entropy and are not conducive to the knowledge distillation process \cite{shen2021label,zhang2022quantifying}.
We provide a theoretical analysis based on the definition of information entropy.

\noindent \textbf{Information Entropy.} 
The entropy of a random variable is the average level of ``information" or ``uncertainty" \cite{shannon1948mathematical} inherent to the variable's possible outcomes. Given a discrete random variable $X$, which takes values in the alphabet $\mathcal{X}$ and is distributed according to $p: \mathcal{X} \to [0,1]:$
\begin{equation}\label{eq2}
H(X)=-\sum_{x\in \mathcal{X}} p(x)\log p(x)=\mathbb{E}[-\log p(X)],
\end{equation}
where $\sum$ denotes the sum over the variable's possible values.
Based on the definition, in the existing process, the teacher's soft label will also change in the existing generation process as ``information" decreases and ``uncertainty" decreases.
Although better distribution similarity between generated and original domains \cite{chen2019data,fang2021contrastive}, we think existing methods ignore the information content of data and the teacher's soft labels.
A trade-off relationship rather than ignoring one of these two helps improve student performance (Tests and analyses are shown in Table \ref{tab3}).

\section{Data-Free Adversarial Robustness Distillation}
\label{sec5}
According to the above analysis, we clarify the motivation and perform simple analytical tests.
In this section, we try to use the adaptive approach to improve the efficiency of knowledge transfer and ensure higher data information while reducing hyperparameter tuning costs.
Firstly, we propose an Interactive Temperature Adjustment (ITA) strategy, which dynamically adjusts the distillation temperature according to the training status of the students in the current training epoch.
The strategy helps students find the appropriate learning objectives for each training epoch.
Secondly, we design an Adaptive Generator Balance (AGB) module to balance similarity and information capacity, avoiding the excessive pursuit of one.
The pipeline is shown in Figure \ref{fig4}.
The generator is trained via the ITA and AGB methods.
Then the student is trained with the ITA strategy.
Besides, the detailed training process is shown in Algorithm \ref{alg1}.

\subsection{Interactive Temperature Adjustment}

Our ITA strategy adjusts the teacher's soft label through the interactive distillation temperature $\tilde{\tau}$ so that the confidence gap between the teacher and student models is kept in a suitable range.
In the generate stage, the generated data should transfer the information of decision boundary from the teacher model to the student model as effectively as possible \cite{heo2019knowledge}.
Unlike previous methods, our generator does not directly synthesize specific data but starts from accessible data for the student to learn.
We maximize the predictions of the student and the teacher to find effective generated data via the adversarial generation loss as:
\begin{equation}\label{eq3}
\mathcal{L}_{adv}=-K\!L(f_t(\hat{\textbf{\textrm{x}}}; \theta_t), f_s(\hat{\textbf{\textrm{x}}}; \theta_s), \tilde{\tau}),
\end{equation}
where $K\!L$ denotes the Kullback-Leibler (KL) divergence loss. 
Then we calculate the teacher's confidence for the generated data and collect the numerical value of the confidence $Con_t$ and the class with the highest confidence $c$ as $Con_t,c = \mathop{\arg\max}f_t(\hat{\textbf{\textrm{x}}})$.
The student's confidence for class $c$ can be directly obtained as $Con_s$. The interactive temperature $\tilde{\tau}$ can be calculated as:
\begin{equation}\label{eq4}
\tilde{\tau}= max\left \{ \frac{1}{bs} \sum_{i=1}^{bs}  \left | Con_t-Con_s \right | \cdot C, 1 \right \}, 
\end{equation}
where $C$ is total number of classes and $bs$ denotes the batch size.
The calculated absolute value represents the difference between the student's and the teacher's prediction in the current training epoch, thus reflecting the current learning situation.
In the early epochs, higher distillation temperatures are set to obtain the generated data that is easier for students to learn.
As the student's prediction gets closer to the teacher's, the temperature drops to synthesize more challenging data.
Notably, ITA can be combined with other generation methods as a plug-and-play strategy.

Similarly, we consider the effectiveness of knowledge transfer in the distillation stage.
As the progress of the students continues to increase the learning difficulty, we define the interactive knowledge distillation loss as:
\begin{equation}\label{eq5}
\mathcal{L}_{K\!D}=\sum_{\hat{\textbf{\textrm{x}}}'\in{\hat{\textbf{\textrm{X}}}'}} KL(f_t({\hat{\textbf{\textrm{x}}}'}; \theta_t), f_s(\hat{\textbf{\textrm{x}}}'; \theta_s), \tilde{\tau}), 
\end{equation}
where $\hat{\textbf{\textrm{x}}}'$ is the adversarial examples
of the generated data $\hat{\textbf{\textrm{x}}}$ \cite{goldblum2020adversarially}.

\subsection{Adaptive Generator Balance}

For generator training objectives, we choose the common training losses (called Vanilla DFARD) to elaborate on our proposed optimization for simplicity and persuasiveness.
The proposed AGB module can adaptively adjust the weight of the losses to balance the domain similarity and data information content according to the current confidence of the teacher (related to information entropy).
For the generator $g$, we combine Equation (\ref{eq1}) and (\ref{eq3}) as:
\begin{equation}\label{eq6}
\mathcal{L}_{gen}=\lambda\cdot\mathcal{L}_{cls}+(1-\lambda)\cdot\mathcal{L}_{adv}, 
\end{equation}
where $\lambda$ is the trade-off parameter.
When the teacher's confidence is too high, the information content of the data may be ignored.
At this time, $\lambda$ adaptively reduces to avoid blindly pursuing similarity.
Specifically, $\lambda$ is calculated as:
\begin{equation}\label{eq7}
\lambda = \frac{1}{C\cdot \frac{1}{bs} \sum_{i=1}^{bs} Con_t}. 
\end{equation}

\noindent The average confidence $ \frac{1}{bs} \sum_{i=1}^{bs} Con_t$ is greater than or equal to the randomly expected value $\frac{1}{C}$ for the dataset with the number of classes $C$.
Therefore, it always satisfies $0<\lambda\le1$.
With the help of AGB, we can increase the amount of information in the generated data while satisfying the similarity, which helps the student's performance.
Simultaneously, we no longer need to try many different weight combinations to test results.
Therefore, our method is more simple and more convenient.

\begin{algorithm}[h]
  \caption{Training process of our Data-Free Adversarial Robustness Distillation}
  \label{alg1}
  \begin{algorithmic}[1]
    \Require
      A pre-trained teacher network $f_{t}$, a generator $g$ with parameter $\theta_g$, a student $f_{s}$ with parameter $\theta_s$, distillation epochs $T$, the iterations of generator $g$ in each epoch $T\!g$, the iterations of student $f_{s}$ in each epoch $T\!s$.

    \State Initialize parameter $\theta_g$ and $\theta_s$
    \For {$i$ in $[1,\dots,T]$}
        \State $//$ \textit{\textbf{Generation stage}}
        \For {$j$ in $[1,\dots,T\!g]$}
        \State Randomly sample noises and labels $(\textbf{\textrm{z}},\textbf{\textrm{y}})$ 
        \State Synthesize training data $\hat{\textbf{\textrm{x}}}=g(\textbf{\textrm{z}},\textbf{\textrm{y}})$
        \State Update generator $g$ through Equation (\ref{eq6})
        \EndFor

        \State $//$ \textit{\textbf{Distillation stage}}
        \For {$j$ in $[1,\dots,T\!s]$}
        \State Synthesize training data $\hat{\textbf{\textrm{x}}}=g(\textbf{\textrm{z}},\textbf{\textrm{y}})$ 
        \State Generate adversarial examples
        $\hat{\textbf{\textrm{x}}}\to\hat{\textbf{\textrm{x}}}'$ 
        \State Distill the student $f_{s}$ through Equation (\ref{eq5})
        \EndFor
    \EndFor
    \Ensure The student $f_{s}$ with adversarial robustness.
    \end{algorithmic}
\end{algorithm}

\begin{table*}[t]
\centering
\setlength{\tabcolsep}{2.6mm}
\scalebox{0.71}{
\begin{tabular}{@{}c|c|ccccccc|ccccccc@{}}
\toprule
\multirow{3}{*}{Model} &
  \multirow{3}{*}{Method} &
  \multicolumn{7}{c|}{CIFAR-10} &
  \multicolumn{7}{c}{CIFAR-100} \\ \cmidrule(l){3-16} 
 &
   &
  \multicolumn{7}{c|}{Attacks Evaluation} &
  \multicolumn{7}{c}{Attacks Evaluation} \\
 &
   &
  Clean &
  FGSM &
  PGD${\rm _{S}}$ &
  PGD${\rm _{T}}$ &
  CW &
  AA &
  \multicolumn{1}{c|}{Average} &
  Clean &
  FGSM &
  PGD${\rm _{S}}$ &
  PGD${\rm _{T}}$ &
  CW &
  AA &
  Average \\ \midrule
\multirow{9}{*}{RN-18} &
  Dream &
  \textbf{68.26} &
  34.76 &
  29.72 &
  31.36 &
  27.96 &
  26.70 &
  \multicolumn{1}{c|}{30.10} &
  22.00 &
  10.18 &
  9.52 &
  9.85 &
  7.11 &
  6.68 &
  8.67 \\
 &
  DeepInv &
  64.53 &
  35.18 &
  31.26 &
  32.49 &
  28.77 &
  27.93 &
  \multicolumn{1}{c|}{31.13} &
  40.91 &
  19.46 &
  17.86 &
  18.68 &
  15.27 &
  14.54 &
  17.16 \\
 &
  DAFL &
  54.98 &
  27.04 &
  24.75 &
  25.87 &
  22.90 &
  22.25 &
  \multicolumn{1}{c|}{24.56} &
  41.67 &
  21.42 &
  20.13 &
  20.81 &
  17.96 &
  17.16 &
  19.50 \\
 &
  DFAD &
  57.58 &
  31.54 &
  29.68 &
  30.65 &
  26.94 &
  26.47 &
  \multicolumn{1}{c|}{29.06} &
  37.57 &
  18.95 &
  17.53 &
  18.14 &
  15.06 &
  14.57 &
  16.85 \\
 &
  ZSKT &
  58.08 &
  31.98 &
  29.94 &
  30.92 &
  27.21 &
  26.68 &
  \multicolumn{1}{c|}{29.35} &
  38.91 &
  20.16 &
  18.78 &
  19.41 &
  16.38 &
  15.52 &
  18.05 \\
 &
  DFQ &
  54.44 &
  26.90 &
  24.63 &
  25.78 &
  22.37 &
  21.57 &
  \multicolumn{1}{c|}{24.25} &
  45.24 &
  22.49 &
  20.78 &
  21.61 &
  18.24 &
  17.38 &
  20.10 \\
 &
  CMI &
  53.28 &
  25.78 &
  23.14 &
  23.97 &
  21.03 &
  20.38 &
  \multicolumn{1}{c|}{22.86} &
  45.04 &
  22.78 &
  21.02 &
  21.90 &
  17.90 &
  16.97 &
  20.11 \\
 &
  Fast &
  61.13 &
  31.40 &
  28.01 &
  29.17 &
  26.26 &
  25.42 &
  \multicolumn{1}{c|}{28.05} &
  36.75 &
  18.66 &
  17.72 &
  18.33 &
  15.57 &
  14.77 &
  17.01 \\
 &
  Ours* &
  65.10 &
  36.36 &
  33.47 &
  34.89 &
  30.79 &
  30.06 &
  \multicolumn{1}{c|}{33.11} &
  45.33 &
  24.08 &
  22.71 &
  23.38 &
  19.84 &
  19.00 &
  21.80 \\
 &
  Ours &
  66.44 &
  \textbf{38.53} &
  \textbf{35.94} &
  \textbf{37.15} &
  \textbf{32.79} &
  \textbf{32.14} &
  \multicolumn{1}{c|}{\textbf{35.31}} &
  \textbf{46.33} &
  \textbf{24.56} &
  \textbf{22.94} &
  \textbf{23.59} &
  \textbf{20.12} &
  \textbf{19.19} &
  \textbf{22.08} \\ \midrule
\multirow{9}{*}{MN-V2} &
  Dream &
  \textbf{64.95} &
  32.03 &
  26.09 &
  27.63 &
  23.83 &
  22.28 &
  \multicolumn{1}{c|}{26.37} &
  18.73 &
  9.78 &
  8.96 &
  9.37 &
  6.93 &
  6.33 &
  8.27 \\
 &
  DeepInv &
  59.53 &
  31.76 &
  28.42 &
  29.74 &
  25.86 &
  24.99 &
  \multicolumn{1}{c|}{28.15} &
  37.75 &
  16.94 &
  15.54 &
  16.19 &
  12.65 &
  11.80 &
  14.62 \\
 &
  DAFL &
  47.53 &
  24.51 &
  21.18 &
  22.09 &
  19.50 &
  18.86 &
  \multicolumn{1}{c|}{21.23} &
  40.46 &
  20.63 &
  19.03 &
  19.78 &
  16.54 &
  15.82 &
  18.36 \\
 &
  DFAD &
  56.13 &
  29.73 &
  26.48 &
  27.64 &
  24.35 &
  24.02 &
  \multicolumn{1}{c|}{26.44} &
  25.41 &
  12.75 &
  11.42 &
  11.95 &
  9.58 &
  9.24 &
  10.99 \\
 &
  ZSKT &
  57.02 &
  30.29 &
  27.07 &
  28.25 &
  24.89 &
  24.40 &
  \multicolumn{1}{c|}{26.98} &
  25.16 &
  12.34 &
  11.36 &
  11.78 &
  9.69 &
  9.16 &
  10.87 \\
 &
  DFQ &
  44.25 &
  21.13 &
  19.14 &
  20.07 &
  16.87 &
  16.20 &
  \multicolumn{1}{c|}{18.68} &
  40.26 &
  19.45 &
  17.74 &
  18.44 &
  15.14 &
  14.35 &
  17.02 \\
 &
  CMI &
  44.53 &
  21.34 &
  19.67 &
  19.97 &
  16.25 &
  15.97 &
  \multicolumn{1}{c|}{18.64} &
  40.23 &
  19.76 &
  17.96 &
  18.56 &
  14.86 &
  14.02 &
  17.03 \\
 &
  Fast &
  54.06 &
  28.23 &
  25.69 &
  26.83 &
  23.18 &
  22.42 &
  \multicolumn{1}{c|}{25.27} &
  38.69 &
  18.58 &
  16.77 &
  17.58 &
  14.62 &
  13.75 &
  16.26 \\
 &
  Ours* &
  59.79 &
  32.25 &
  29.25 &
  30.24 &
  26.18 &
  25.56 &
  \multicolumn{1}{c|}{28.70} &
  40.94 &
  21.47 &
  20.18 &
  20.89 &
  17.60 &
  16.82 &
  19.39 \\
 &
  Ours &
  61.16 &
  \textbf{34.46} &
  \textbf{31.66} &
  \textbf{32.80} &
  \textbf{28.40} &
  \textbf{27.90} &
  \multicolumn{1}{c|}{\textbf{31.04}} &
  \textbf{41.78} &
  \textbf{22.04} &
  \textbf{20.84} &
  \textbf{21.68} &
  \textbf{17.93} &
  \textbf{17.04} &
  \textbf{19.91} \\ \bottomrule
\end{tabular}
}
\vspace{-0.05cm}
\caption{Adversarial robustness accuracy (\%) on CIFAR-10 and CIFAR-100. The maximum adversarial perturbation $\epsilon$ is 8/255. 
\textbf{Bold} numbers denote the best results.
\textit{Average} indicates the average value of the robustness test, which does not include the clean accuracy.
``Ours*" means training the generator as Equation (\ref{eq6}).
``Ours" means the complete method as Algorithm \ref{alg1}.
}
\label{tab1}
\end{table*}

\begin{table*}[t]
\centering
\setlength{\tabcolsep}{2.7mm}
\scalebox{0.76}{
\begin{tabular}{@{}ccccccccc@{}}
\toprule
Method    & Dream & DeepInv & DAFL & ZSKT & DFQ & CMI & Fast & Ours \\ \midrule
CIFAR-10  &   29.17$h*m$    &  24.70$h*nm$       &  4.04$h*nm$    &  3.03$h*n$     &  4.09$h*nm$    &  48.69$h*nm$    & 6.10$h*nm$      &   3.28$h$    \\
CIFAR-100 &   125.14$h*m$     &  101.28$h*nm$       &  13.43$h*nm$     &  5.62$h*n$     & 13.11$h*nm$     & 77.99$h*nm$     & 12.16$h*nm$      & 5.85$h$      \\ \bottomrule
\end{tabular}
}
\vspace{-0.05cm}
\caption{The synthesis time of various data-free generation methods. We test the specific GPU time on a single RTX 3090 for the entire generation process. $h$ is short for hours, $n$ denotes the distillation temperature hyperparameter tuning times, and $m$ denotes the generator loss weights tuning times.}
\label{tab2}
\end{table*}

\section{Experiments}

\subsection{Experimental Setup}

\textbf{Dataset and Model.} 
We evaluate the proposed DFARD method on 32$\times$32 CIFAR-10 and CIFAR-100 datasets \cite{krizhevsky2009learning}, which are the most commonly used datasets for testing adversarial robustness.
For a fair comparison, we use the same pre-trained WideResNet (WRN) teacher models with \cite{zi2021revisiting}.
Furthermore, we evaluate all methods using the student with ResNet-18 (RN-18) \cite{he2016deep} and MobileNet-V2 (MN-V2) \cite{sandler2018mobilenetv2} following existing ARD methods \cite{zhu2022reliable,zi2021revisiting}.

\noindent \textbf{Baselines.} 
We compare our optimized method with different data-free generation methods, including Dream \cite{bhardwaj2019dream}, DeepInv \cite{yin2020dreaming}, DAFL \cite{chen2019data}, DFAD \cite{fang2019data}, ZSKT \cite{micaelli2019zero}, DFQ \cite{choi2020data}, CMI \cite{fang2021contrastive}, and Fast \cite{fang2022up}.
We use the same PGD-attack to generate the adversarial examples to train the student for all baseline methods.
For a fair comparison, the distillation process uses the same training loss as ARD \cite{goldblum2020adversarially}.

\noindent \textbf{Implementation details.}
Our proposed method and all others are implemented in PyTorch.
All models are trained on RTX 3090 GPUs \cite{paszke2019pytorch}.
The students are trained via SGD optimizer with cosine annealing learning rate with an initial value of 0.05, momentum of 0.9, and weight decay of 1e-4.
The generators are trained via Adam optimizer with a learning rate of 1e-3, $\beta_1$ of 0.5, $\beta_2$ of 0.999.
The distillation batch size and the synthesis batch size are both 256.
The distillation epochs $T$ is 200, the iterations of generator $T\!g$ is 1, and the iterations of student $T\!s$ is 5.
Both the student model and the generator are randomly initialized.
A 10-step PGD (PGD-10) with a random start size of 0.001 and step size of $2/255$ is used to generate adversarial samples.
The perturbation bounds are set to $L_\infty $ norm $\epsilon=8/255$.

\noindent \textbf{Attack Evaluation.}
We evaluate the adversarial robustness with five adversarial attacks: FGSM \cite{goodfellow2014explaining}, PGD${\rm _{S}}$ \cite{madry2017towards}, PGD${\rm _{T}}$ \cite{zhang2019theoretically}, CW$_\infty$ \cite{carlini2017towards} and AutoAttack (AA) \cite{croce2020reliable}.
These methods are the most commonly used for adversarial robustness evaluation.
The maximum perturbation is set as $\epsilon=8/255$.
The perturbation steps for PGD${\rm _{S}}$, PGD${\rm _{T}}$ and CW$_\infty$ are 20.
In addition, we test the accuracy of the models in normal conditions without adversarial attacks (Clean).

\subsection{Comparison with Other Methods}

To compare the effects of various data-free generation methods, we set the same distillation process as ARD \cite{goldblum2020adversarially}.
Therefore, the difference only lies in the generator loss function of these methods.
We select and report the best checkpoint of all methods among all epochs.
The best checkpoints are based on the adversarial robustness performance against PGD${\rm _{T}}$ attack.
For the computational costs, we compare the synthesis time on the generation stage of different generation methods.

\noindent \textbf{Performance Comparison.} 
The robustness performances of our and other baseline methods are shown in Table \ref{tab1}. 
Our generation method (\textbf{Ours*}) achieves better adversarial robustness performance in all baselines.
The results demonstrate that our interactive and adaptive approach can be more effective for the challenging DFARD task.
For different backbone and dataset combinations, our method improves the average adversarial robustness by 1.98\%, 0.55\%, 1.69\%, and 1.03\%, respectively, compared to other best results.

Notably, our method maintains the most stable performance in various settings, while others may perform poorly in some settings.
We consider that one reason is our interactive learning objective, which helps to improve students' versatility in different settings.
Specifically, Dream \cite{bhardwaj2019dream} inverts enough data for normal ARD.
However, these data might not be suitable for student learning as the data comes exclusively from teachers.
DeepInv \cite{yin2020dreaming} and CMI \cite{fang2021contrastive} excessively pursue distribution similarity between generated and original domains ignoring the information content of data.
Fast \cite{fang2022up} uses a feature-sharing method, but the lack of rich new features in complex datasets leads to performance degradation.
In contrast, some early methods (DAFL \cite{chen2019data}, DFAD \cite{fang2019data}, ZSKT \cite{micaelli2019zero} and DFQ \cite{choi2020data}) are more stable and effective, but these methods keep the same teacher predictions throughout the generation process.
Therefore, their learning objectives may not meet every epoch for the randomly initialized students.
Good results are often inseparable from multiple hyperparameter tuning.

\begin{figure*}[th]
	\centering
	\includegraphics[scale=0.81]{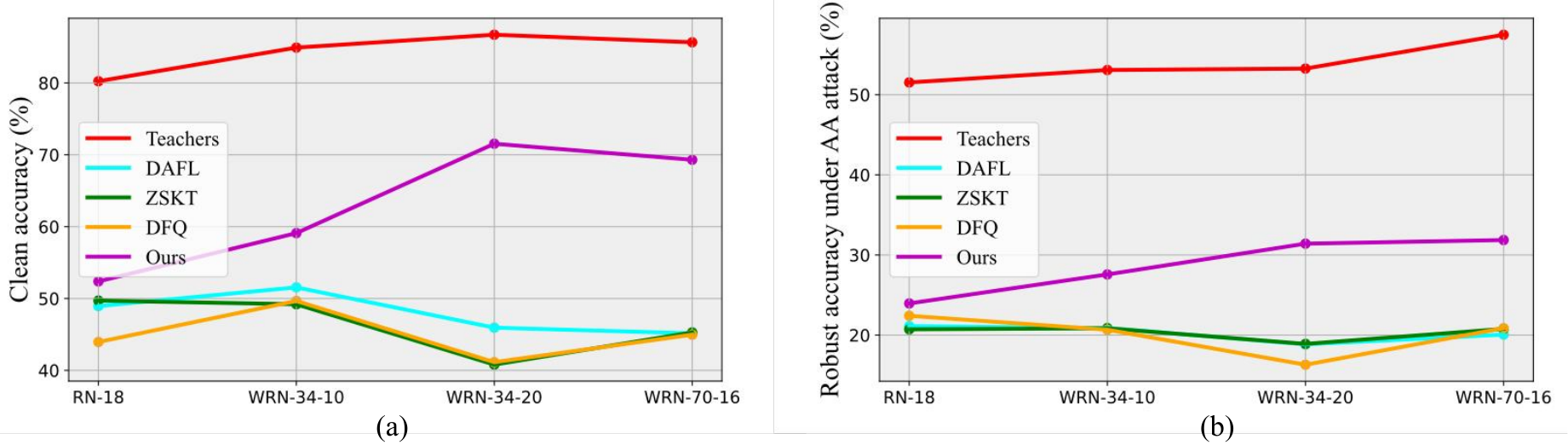}
    \vspace{-0.15cm}
	\caption{Performance of RN-18 students trained with different teachers. Students train 100 epochs for 4 generative methods. (a) shows clean accuracy, and (b) shows robust accuracy under AA attack.}
	\label{fig5}
    \vspace{-0.1cm}
\end{figure*}

\noindent \textbf{Generation Computational Costs.} 
In addition, we also compare the overall generation computational costs of all methods while considering the hyperparameter tuning calculation costs.
The results are shown in Table \ref{tab2}.
Other methods must test multiple sets of temperature parameters (denoted by $n$) or trade-offs between multiple generator losses (denoted by $m$).
Notably, some methods require significantly higher weights tuning times $m$ when using four or more generator losses, \textit{e.g.}, Dream, DeepInv, DFQ, and CMI.
Thanks to fully adaptive parameters, our generator's computational costs are significantly lower than most other methods.
In summary, our method has the most stable performance while maintaining the advantage of significantly lower generation cost.
Therefore, our method is simple, reliable, and convenient.

\subsection{Adaptability for Different Teachers}

The method's adaptability is important to reduce reliance on customized teachers \cite{zi2021revisiting}.
To compare the adaptability for different teachers, we select four teachers (RN-18, WRN-34-10, WRN-34-20, WRN-70-16 \cite{croce2021robustbench}) to train the RN-18 student on CIFAR-10.
The results are shown in Figure~\ref{fig5}.
For the other methods, we find the robust saturation phenomenon.
Due to the capacity gap between the teacher and student, the learning objectives provided by larger-scale teachers may not be suitable for small students to learn.
However, our method is less susceptible to the gap due to the interactive learning objectives.
Our proposed easy-to-hard process alleviates the saturation and has more vital adaptability for teachers with different capacities.

\begin{table}[t]
\centering
\setlength{\tabcolsep}{1.8mm}
\scalebox{0.67}{
\begin{tabular}{@{}cc|ccccccc@{}}
\toprule
\multirow{2}{*}{ID} & \multirow{2}{*}{Settings}      & \multicolumn{7}{c}{Attacks Evaluation}                                      \\
                    &                                & Clean & FGSM  & PGD${\rm _{S}}$ & PGD${\rm _{T}}$ & CW    & AA    & Average \\ \midrule
(1)                 & Vanilla DFARD & 58.35 & 30.68 & 28.59           & 29.75           & 24.67 & 23.95 & 27.53   \\
(2)                 & w/ ITA (G)           & 59.16 & 31.05 & 28.59           & 29.90           & 24.87 & 24.21 & 27.72   \\
(3)                 & w/ ITA (G+D)        & 60.36 & 33.09 & 30.97           & 32.02           & 26.79 & 26.24 & 29.82   \\
(4)                 & w/ AGB                 & 64.41 & 36.33 & 33.53           & 34.91           & 30.57 & 29.92 & 33.05   \\
(5)                 & w/ AGB w/ ITA (G)              & 65.10 & 36.36 & 33.47           & 34.89           & 30.79 & 30.06 & 33.11   \\
(6)                 & \textbf{Ours}                           & \textbf{66.44} & \textbf{38.53} & \textbf{35.94}           & \textbf{37.15}           & \textbf{32.79} & \textbf{32.14} & \textbf{35.31}   \\ \bottomrule
\end{tabular}
}
\vspace{-0.05cm}
\caption{Ablation study on CIFAR-10.
For vanilla DFARD, we choose the best hyperparameters (the distillation temperature $\tau = 3$, the loss weight $\lambda = 0.3$).
`G’ means that ITA is applied only in the generation stage, and `G+D’ means that ITA is applied in both the generation and distillation stages.}
\label{tab3}
\vspace{-0.1cm}
\end{table}

\subsection{Ablation Study}

\textbf{Impact of Interactive Temperature Adjustment.}
To thoroughly verify the effectiveness of the proposed Interactive Temperature Adjustment (ITA) strategy, we test it in both the generation and distillation stages.
As shown in Table~\ref{tab3}(1-3) and (4-6), compared with the best-fixed temperature parameters (w/o ITA), students' performance improves when ITA is applied in the generation stage.
Further, when the interactive temperature is deployed in the both generation and distillation stages, the performance improves again.
It is worth noting that Table~\ref{tab3} does not reflect hyperparameter tuning time.
The best temperature comes from multiple experimental tests.
Even so, the student's feedback can dynamically adjust the difficulty of knowledge transfer to improve student performance, which verifies the effectiveness of ITA.

\begin{table}[t]
\centering
\setlength{\tabcolsep}{3.5mm}
\scalebox{0.75}{
\begin{tabular}{@{}lccc@{}}
\toprule
\multicolumn{1}{c}{Method} & Clean         & CW            & AA            \\ \midrule
DAFL\,+\,ITA                   & 56.26 \small{\textcolor[rgb]{0.4,0.4,0.4}{(+1.28)}} & 24.78 \small{\textcolor[rgb]{0.4,0.4,0.4}{(+1.88)}} & 24.36 \small{\textcolor[rgb]{0.4,0.4,0.4}{(+2.11)}} \\
DFQ\;\,+\;\,ITA                    & 55.79 \small{\textcolor[rgb]{0.4,0.4,0.4}{(+1.35)}} & 24.12 \small{\textcolor[rgb]{0.4,0.4,0.4}{(+1.75)}} & 23.74 \small{\textcolor[rgb]{0.4,0.4,0.4}{(+2.17)}} \\
ZSKT\,+\,ITA                   & 59.93 \small{\textcolor[rgb]{0.4,0.4,0.4}{(+1.85)}} & 29.03 \small{\textcolor[rgb]{0.4,0.4,0.4}{(+1.82)}} & 28.58 \small{\textcolor[rgb]{0.4,0.4,0.4}{(+1.90)}} \\ \bottomrule
\end{tabular}
}
\vspace{-0.05cm}
\caption{Other methods with our proposed ITA.}
\label{tab4}
\vspace{-0.1cm}
\end{table}

\noindent \textbf{Other Methods with the ITA Strategy.}
Further, we combine the proposed plug-and-play Interactive Temperature Adjustment (ITA) strategy with three other methods for both the generation and distillation stages.
Experiments are carried out on CIFAR-10 with the RN-18 student.
The results are shown in Table \ref{tab4}.
Compared with the baseline performance in Table~\ref{tab1}, for the three methods, both clean and robust accuracy are significantly improved, which proves that our ITA strategy promotes student performance through an easy-to-hard knowledge transfer process.

\noindent \textbf{Impact of Adaptive Generator Balance.}
Further, we evaluate the effectiveness of the proposed AGB module.
The results are shown in Table~\ref{tab3} (4-6).
Compared with the best-fixed $\lambda$, our AGB module significantly improves student performance for both clean and robust accuracy.
At the same time, our adaptive approach omits the weight-tuning costs.

\section{Conclusion}

This paper proposes a novel task named Data-Free Adversarial Robustness Distillation (DFARD) to deal with realistic scenes with higher security levels.
We demonstrate that the task is more challenging than existing tasks, making combining previous methods less effective.
Due to less information available, we try to optimize the new task by improving knowledge transfer efficiency and maintaining higher data information.
Our method is simple yet effective, achieving stable performance while maintaining the advantage of significantly lower generation cost.
We believe the proposed technique helps apply deep learning techniques to real scenes.

\section{Acknowledgments}
This work is supported by Shanghai Municipal Science and Technology Major Project (No.2021SHZDZX0103), the Shanghai Engineering Research Center of AI \& Robotics, Fudan University, China, and the Engineering Research Center of AI \& Robotics, Ministry of Education, China.

\bigskip

\bibliography{aaai24}

\newpage

\appendix

\noindent {\LARGE\textbf{Appendix}}

\section{The Proofs for why DFARD is challenging.}

To illustrate the challenge of DFARD, we first introduce the Lipschitzness to analyze the knowledge transfer process.

\subsection{Lipschitzness.} A mapping $f:\mathbb{R}^n \to \mathbb{R}^m$ is said to be L-Lipschitz continuous \textit{w.r.t.} norm
$\left | \; \cdot \; \right | $ if for any pair of inputs ${x}_1, {x}_2 \in \mathbb{R}^n$,
\begin{equation}\label{eq8}
\left | f({x}_1)-f({x}_2) \right | \le L \left |  {x}_1- {x}_2  \right |.  
\end{equation}
For different deep learning tasks, the Lipschitz coefficients $L$ are different. 
According to \cite{croce2020reliable} and \cite{zhang2022rethinking}, the Lipschitz coefficient $L$ in the model with adversarial robustness is smaller than the normal non-robust model.

\subsection{Standard Lipschitz Networks.} As with the teacher-student network structures used for all experiments in this paper, we assume that the networks in the knowledge transfer process satisfy the Standard Lipschitz Networks, which are formed by affine layers (\textit{e.g.}, the fully connected or convolutional layer) and element-wise activation functions.
For a multi-layer teacher network, the storage form of knowledge is expressed as:
\begin{equation}\label{eq9}
\small
f(x_1) = \sigma^{(l)}(\mathrm{W}^{(l)} \sigma^{(l-1)}(\mathrm{W}^{(l-1)}\cdots \sigma^{(1)}(\mathrm{W}^{(1)}{x}_1)\cdots), 
\end{equation}
\begin{equation}\label{eq10}
\small
f(x_2) = \sigma^{(l)}(\mathrm{W}^{(l)} \sigma^{(l-1)}(\mathrm{W}^{(l-1)}\cdots \sigma^{(1)}(\mathrm{W}^{(1)}({x}_1+ \Delta))\cdots), 
\end{equation}
where $l \in [M] $. $M$ is the number of layers.
$\sigma$ is the identity function. $\mathrm{W}$ is the weight parameter.
The bias parameter $b$ of each layer is omitted here for convenience.
The input ${x}$ meets ${x} \in ({x}_1,\cdots,{x}_{bs})$, \textit{i.e.}, ${x} \in ({x}_1,{x}_1+\Delta_1,\cdots,{x}_1+\Delta_{bs-1})$.
$\Delta$ represents the difference among different samples.
For a standard lipschitz network $f$, we combine Formulas (\ref{eq8}), (\ref{eq9}), (\ref{eq10}) to get $ f(x_2) \le f(x_1)+ L\sigma_t \mathrm{W}_t(\Delta_1) $.
In the same way, for the \textit{i}-th sample, the prediction of the network satisfies:
\begin{equation}\label{eq11}
f(x_i) \le f(x_1)+ L\sigma_t \mathrm{W}_t(\Delta_{i-1}).
\end{equation}

\begin{table*}[t]
\centering
\caption{Teacher models used in our experiments and their performance.}
\setlength{\tabcolsep}{3.5mm}
\scalebox{0.85}{
\begin{tabular}{@{}ccccccccc@{}}
\toprule[1pt]
                Setting       & Dataset   & Teacher          & Clean   & FGSM    & PGD${\rm _{S}}$ & PGD${\rm _{T}}$ & CW      & AA      \\ \midrule
\multirow{2}{*}{Default} & CIFAR-10 & WideResNet-34-10 & 84.92\% & 60.87\% & 55.33\% & 56.61\% & 53.98\% & 53.08\% \\
                       & CIFAR-100 & WideResNet-70-16 & 60.86\% & 35.68\% & 33.56\%         & 33.99\%         & 31.05\% & 30.03\% \\ \midrule
\multirow{4}{*}{Other} & CIFAR-10  & ResNet-18        & 80.24\%   & 59.76\%   & 56.27\%           & 57.27\%           & 52.05\%   & 51.53\%   \\

& CIFAR-10 & WideResNet-34-10 & 84.92\% & 60.87\% & 55.33\% & 56.61\% & 53.98\% & 53.08\%  \\
                       & CIFAR-10  & WideResNet-34-20 &   88.70\%      &   63.26\%      &    55.91\%             & 57.74\%                &   54.76\%      &  53.26\%       \\
                       & CIFAR-10  & WideResNet-70-16 & 85.66\%   & 64.80\%   & 61.09\%           & 62.39\%           & 58.52\%   & 57.48\%   \\ \bottomrule[1pt]
\end{tabular}
}
\label{tab5}
\end{table*}

\subsection{Knowledge Transfer Process.} 
In the process of knowledge distillation, the knowledge transfer process can be expressed as:
\begin{equation}\label{eq12}
\mathcal{L}_{K\!D}(\mathrm{W}_s, \mathrm{W}_t) = \lambda \cdot \frac{1}{bs} \sum_{i=1}^{bs} D(\sigma_s \mathrm{W}_s({x}_i) - \sigma_t \mathrm{W}_t({x}_i)), 
\end{equation}
where $D$ is a distance constraint function. $bs$ denotes the batch size. $\lambda$ is related to the learning rate.
We denote $K$ as the knowledge transfer information. In the general knowledge transfer process (\textit{i.e.}, regardless of the unsupervised methods), $K$ relies almost entirely on the label representation provided by the teacher. Considering the loss of information, $K$ is not greater than the maximum information the teacher can provide:
\begin{equation}\label{eq13}
K \le \lambda \cdot \frac{1}{bs} \sum_{i=1}^{bs} \sigma_t \mathrm{W}_t({x}_i).
\end{equation}

Combining with Formula (\ref{eq11}), we can get the knowledge that students can effectively accept satisfies the following:
\begin{equation}\label{eq14}
\begin{split}
K \le \lambda \cdot \frac{1}{bs} \sum_{i=1}^{bs} \sigma_t \mathrm{W}_t({x}_i) = \lambda \cdot \frac{1}{bs} \sum_{i=1}^{bs} f({x}_i) \\
\le \lambda \cdot f({x}_1)+L\sigma_t \mathrm{W}_t \sum_{i=1}^{bs-1} \Delta_i, 
\end{split}
\end{equation}
Based on Formula (\ref{eq14}), we get an upper bound about the knowledge transfer process.
Since the coefficient $L$ in the robustness task is much smaller than the normal distillation training, the information upper bound of the robustness task is smaller than the normal training, \textit{i.e.}, DFARD has a lower information upper limit than KD in the knowledge transfer process.
Simultaneously, the synthetic samples of current data-free generation methods are far less diverse than the original data domain \cite{ye2020data}.
The $\Delta$ in the data-free task is much smaller than $\Delta$ in the data-based task.
Based on this, the upper bound of knowledge transfer information in data-free tasks is also smaller than that in data-based tasks, \textit{i.e.}, DFARD has a lower information upper limit than ARD in the knowledge transfer process.

Therefore, in the DFARD task, more knowledge is needed.
In this paper, we aim to improve the efficiency of knowledge transfer and ensure higher data information to tackle the challenging task.

\section{The pre-trained teacher models used in the experimental part}

For a fair comparison with other SOTA methods, we use the same pre-trained WideResNet (WRN) \cite{zagoruyko2016wide} teacher models (Table~\ref{tab5} Default) with RSLAD \cite{zi2021revisiting}.
In further analysis, we use various teacher models \cite{croce2021robustbench} (Table~\ref{tab5} Other) to test the student’s adaptability for teachers with different capacities.
The capacity of the model increases gradually from top to bottom.

\section{Experimental settings for other baseline methods.}

\textbf{Dream.}
Dream~\cite{bhardwaj2019dream} is a model inversion method.
We keep the inversion loss from the original as: $\mathcal{L}_{ce}=CE(f_{t}(\hat{{{x}}}), {{y}})$.
The same hyperparameters as the original paper were used, including: 50 PCs per cluster, generated a total of 50000 synthetic images. We use the Adam optimizer with a learning rate of 0.05, $\beta_1$ of 0.9, $\beta_2$ of 0.999, iterations of 500. 

\noindent \textbf{DeepInv.}
DeepInv~\cite{yin2020dreaming} is a model inversion method that combines prior knowledge and adversarial training.
We keep the inversion loss from the original as: $\mathcal{L}=\alpha_{tv}\mathcal{R}_{tv}+ \alpha_{l2} \mathcal{R}_{l2} + \alpha_{f} \mathcal{R}_{feature} + \alpha_{c}\mathcal{R}_{compete}$.
We set $\alpha_{tv}=2.5e-5$ , $\alpha_{l2}=3e-8$, $\alpha_{f}=0.1$ and $\alpha_{c}=10$, which are the same as the original setting.
Besides, we set the number of iterations as 2000, and use Adam for optimization with a learning rate of 0.05.

\noindent \textbf{DAFL.}
DAFL~\cite{chen2019data} is a data-free generation method.
We keep the generator loss from the original as: $\mathcal{L}_{total}=\mathcal{L}_{oh}+ \alpha\mathcal{L}_{a}+ \beta\mathcal{L}_{ie} $.
Following the original settings, we set $\alpha=1e-3$, $\beta=20$. We use SGD with the weight decay of 5e-4, the momentum of 0.9, and the initial learning rate set as 0.1.

\noindent \textbf{DFAD.}
DFAD~\cite{fang2019data} is a data-free adversarial generation method.
We keep the generator loss from the original as:
$\mathcal{L}_{total}=\mathcal{L}_{L1}$.
We use SGD with the weight decay of 5e-4, the momentum of 0.9, and the initial learning rate set as 0.1.

\noindent \textbf{ZSKT.}
ZSKT~\cite{micaelli2019zero} is a simple and effective data-free adversarial generation method.
We keep the generator loss from the original as: $ \mathcal{L} = - \mathcal{L}_{adv}$.
The iteration for the generator is 1, and the iteration for the student is 10.
The generator uses Adam with an initial learning rate of 1e-3.
The optimizer of the student is the same as the DAFL.

\noindent \textbf{DFQ.}
DFQ~\cite{choi2020data} is also a data-free adversarial distillation method. 
We use the loss as: $\mathcal{L}_{total}=\mathcal{L}_{oh}+ \alpha\mathcal{L}_{bn}+ \beta\mathcal{L}_{balance} $.
We set the $ \alpha$ as 1, the $ \beta$ as 20.
For simple calculation and higher performance, we use the ResNet-34 pre-trained network in the original text when calculating $\mathcal{L}_{bn}$ and use the robust teacher network as other methods when calculating the other two.
The generator uses the Adam optimizer with a momentum of 0.5 and a learning rate of 1e-3.
The student uses the SGD optimizer with a momentum of 0.9 and a learning rate of 0.1.

\noindent \textbf{CMI.}
CMI~\cite{fang2021contrastive} is a model inversion method with contrastive learning.
We keep the generator loss from the original as: $ \mathcal{L} = \alpha\mathcal{L}_{bn}+ \beta\mathcal{L}_{cls} + \gamma\mathcal{L}_{adv} + \delta\mathcal{L}_{cr}$.
We set $\alpha=1$, $\beta=0.5 $, $\gamma=0.5 $, and $\delta=0.8$. The calculation of $ \mathcal{L}_{bn}$ is the same as the DFQ.
We use the Adam Optimizer with a learning rate of 1e-3 to update the generator and the SGD optimizer with a momentum of 0.9 and a learning rate of 0.1 for student training.

\noindent \textbf{Fast.}
Fast \cite{fang2022up} is a fast data-free generation method via feature sharing.
We keep the generator loss from the original as: $ \mathcal{L} = \alpha\mathcal{L}_{cls} + \beta\mathcal{L}_{adv} + \gamma\mathcal{L}_{feat}$.
We set $\alpha=0.4$, $\beta=1.1 $, and $\gamma=10$, which are the same as the original settings.
The optimizer of the generator and student is the same as the CMI.

\section{The generator architecture.}

For a fair comparison, we reproduce all previous methods using the same generator architecture, the default architecture for most of them.
The detailed architecture information is shown in Table \ref{tab_app}.

\begin{table}[h]
\centering
\caption{The generator architecture in this paper. $C$ represents the number of classes.}
\scalebox{1}{
\begin{tabular}{@{}c@{}}
\toprule
Input: $z\in\mathbb{R}^{1024}$, random labels $y\in \mathbb{R}^C $ \\ \midrule
Linear (1024)$\to 8\times8\times256$ \\ \midrule
Reshape, BN \\ \midrule
Upsample $\times$2 \\ \midrule
3$\times$3 Conv 256 $\to$128, BN, LeakyReLU \\ \midrule
Upsample $\times$2 \\ \midrule
3$\times$3 Conv 128 $\to$64, BN, LeakyReLU \\ \midrule
3$\times$3 Conv 64 $\to$3, Sigmoid \\ \midrule
Output: a batch of synthetic data \\ \bottomrule
\end{tabular}
}
\label{tab_app}
\end{table}

\section{More analyses and experimental evidence.}

\subsection{The significance of small model adversarial robustness.}
Paying attention to the versatility of small models is an important part of deploying advanced technologies into practical scenarios.
We want to clarify some facts: 
\textbf{1)} there is a lot of excellent peer-reviewed work exploring the adversarial robustness of small models. 
\textbf{2)} small models are widely deployed on edge devices, are more oriented to the crowd and are more likely to be attacked, so it is meaningful to explore the adversarial robustness.
\textbf{3)} Compared with ARD task, one advantage of DFARD is that it can deal with the situation where the original data is not available due to data privacy, which improves the scope of the application of ARD.

\subsection{The computation overhead of appropriate hyperparameter selection and DFARD task.}
The adaptive parameters mentioned in our paper can be obtained through a few matrix operations. The computational overhead is almost negligible compared to the training overhead of the generator and student models. The results of the specific GPU time on a single RTX 3090 are as follows (One iteration under a mini-batch):
Hyperparameters selection: 0.1828ms; Generator training: 875.1895ms; Student training: 8240.0019ms.
Our hyperparameter calculations (Equations (4) and (7) of the main body) add about 0.002\% additional computational overhead (about 1.43 seconds). 
In contrast, an ablation experiment with manually set hyperparameters, like other methods, requires 200 epoch training (about 20 hours).

In addition, compared with ARD task, the additional overhead of DFARD mainly focuses on introducing the generator and its parameter updates. 
Time calculation overhead (one iteration): ARD 8025.0019ms, DFARD 9115.1914ms (+13.58\%). Memory usage (the batch size equals 256): ARD 9071MB, DFARD 9575MB (+5.56\%).

\begin{table}[th]
\centering
\vspace{-0.1cm}
\caption{The effect of interactive or fixed generator temperature and adaptive or fixed generator loss weights on adversarial robustness.
\textbf{ITA} represents our interactive strategy. \textbf{AGB} represents the adaptive generator balance method.}
\setlength{\tabcolsep}{2.5mm}
\scalebox{0.7}{
\begin{tabular}{@{}cc|ccccccc@{}}
\toprule
\multicolumn{2}{c|}{\multirow{2}{*}{Settings}} &
  \multicolumn{7}{c}{Attacks Evaluation} \\
\multicolumn{2}{c|}{} &
  Clean &
  FGSM &
  PGD${\rm _{S}}$ &
  PGD${\rm _{T}}$ &
  CW &
  AA &
  Average \\ \midrule
\multicolumn{1}{c|}{\multirow{6}{*}{T}} &
  1 &
  42.33 &
  21.89 &
  20.48 &
  21.40 &
  18.32 &
  17.76 &
  19.97 \\
\multicolumn{1}{c|}{} &
  3 &
  42.64 &
  22.15 &
  20.84 &
  21.95 &
  18.95 &
  18.39 &
  20.46 \\
\multicolumn{1}{c|}{} &
  5 &
  42.43 &
  21.74 &
  20.44 &
  21.14 &
  18.63 &
  18.03 &
  20.00 \\
\multicolumn{1}{c|}{} &
  7 &
  41.11 &
  20.61 &
  19.12 &
  19.92 &
  17.70 &
  17.11 &
  18.89 \\
\multicolumn{1}{c|}{} &
  9 &
  41.42 &
  19.81 &
  18.20 &
  18.97 &
  16.35 &
  15.69 &
  17.80 \\
\multicolumn{1}{c|}{} &
  \textbf{ITA} &
  \textbf{43.52} &
  \textbf{22.65} &
  \textbf{21.45} &
  \textbf{22.45} &
  \textbf{18.95} &
  \textbf{18.42} &
  \textbf{20.78} \\ \midrule
\multicolumn{1}{c|}{\multirow{6}{*}{$\lambda$}} &
  0.1 &
  56.41 &
  28.06 &
  25.53 &
  26.63 &
  23.10 &
  22.47 &
  25.16 \\
\multicolumn{1}{c|}{} &
  0.3 &
  59.16 &
  31.05 &
  28.59 &
  29.90 &
  24.87 &
  24.21 &
  27.72 \\
\multicolumn{1}{c|}{} &
  0.5 &
  31.66 &
  13.83 &
  13.04 &
  13.73 &
  9.07 &
  8.59 &
  11.65 \\
\multicolumn{1}{c|}{} &
  0.7 &
  19.35 &
  11.45 &
  10.68 &
  11.28 &
  9.15 &
  8.96 &
  10.30 \\
\multicolumn{1}{c|}{} &
  0.9 &
  10.87 &
  9.58 &
  9.42 &
  9.44 &
  8.77 &
  8.78 &
  9.20 \\
\multicolumn{1}{c|}{} &
  \textbf{AGB} &
  \textbf{65.10} &
  \textbf{36.36} &
  \textbf{33.47} &
  \textbf{34.89} &
  \textbf{30.79} &
  \textbf{30.06} &
  \textbf{33.11} \\ \bottomrule
\end{tabular}
}
\label{tab7}
\end{table}

\subsection{Fixed hyperparameters vs. our adaptive parameters.}
We test the ResNet-18 students on CIFAR-10 with the WRN-34-10 teacher.
\textbf{1)} Table~\ref{tab7} Top shows the results of temperature hyperparameters with the best checkpoint (the first 50 epochs). 
Different temperatures affect the students' adversarial robustness.
Our interactive temperature outputs the fixed and step temperatures and is more likely to avoid parameter tuning costs.
These results certify that using fixed temperature is less efficient for the knowledge transfer process, while the student’s feedback can significantly optimize this process. 
\textbf{2)} Table~\ref{tab7} Bottom shows the results of weights hyperparameters for 200 epochs.
When $\lambda$ is used as a fixed hyperparameter, its size significantly impacts the results, which usually require tuning to get the optimal value. 
The tuning incurs additional computational costs. 
Our adaptive approach omits the above process while significantly improving the baseline performance.

\bigskip

\end{document}